\documentclass[letterpaper, 10 pt, conference]{ieeeconf}
\usepackage{amsmath,amsfonts}
\usepackage{array}
\usepackage{enumerate}
\usepackage[utf8]{inputenc}
\usepackage[T1]{fontenc}
\usepackage{amsfonts}
\usepackage{amssymb}
\usepackage{tabularx}
\makeatletter
\let\NAT@parse\undefined
\makeatother
\usepackage{hyperref}
\hypersetup{
    colorlinks=true,
    linkcolor=blue,
    filecolor=magenta,      
    urlcolor=cyan,
    }
\IEEEoverridecommandlockouts 
\usepackage{algorithm}  
\usepackage{algpseudocode}  
\usepackage{amsmath}  
\usepackage{breqn}
\usepackage[caption=false]{subfig}
\usepackage{textcomp}
\usepackage{stfloats}
\usepackage{url}
\usepackage{verbatim}
\usepackage{graphicx}
\usepackage{cite}
\usepackage{color}
\usepackage{colortbl}
\let\labelindent\relax
\usepackage{enumitem}
\usepackage{multirow}
\usepackage{diagbox}
\usepackage{booktabs}

\title{\LARGE \bf Adaptive Task Allocation in Multi-Human Multi-Robot Teams under Team Heterogeneity and Dynamic Information Uncertainty}

\author{Ziqin Yuan$^{1}\dag$, Ruiqi Wang$^{1}\dag$, Taehyeon Kim$^{1}$, Dezhong Zhao$^{1,2}$, Ike Obi$^{1}$, and Byung-Cheol Min$^{1}$
\thanks{ $\dag$ Equal Contribution}
\thanks{$^{1}$SMART Laboratory, Department of Computer and Information Technology, Purdue University, West Lafayette, IN, USA. {\tt\small{[yuan460, wang5357, kim4435, obii, minb]@purdue.edu}.}}
\thanks{$^{2}$College of Mechanical and Electrical Engineering, Beijing University of Chemical Technology, Beijing, China. \tt\small{DZ\_Zhao@buct.edu.cn}.}
\thanks{This paper is based on research supported by the National Science Foundation (NSF) under Grant No. IIS-1846221.}}

\begin{document}
\setlength{\abovedisplayskip}{1pt} 
\setlength{\belowdisplayskip}{1pt} 

\maketitle

\begin{abstract}
Task allocation in multi-human multi-robot (MH-MR) teams presents significant challenges due to the inherent heterogeneity of team members, the dynamics of task execution, and the information uncertainty of operational states. Existing approaches often fail to address these challenges simultaneously, resulting in suboptimal performance. To tackle this, we propose ATA-HRL, an adaptive task allocation framework using hierarchical reinforcement learning (HRL), which incorporates initial task allocation (ITA) that leverages team heterogeneity and conditional task reallocation in response to dynamic operational states. Additionally, we introduce an auxiliary state representation learning task to manage information uncertainty and enhance task execution. Through an extensive case study in large-scale environmental monitoring tasks, we demonstrate the benefits of our approach. More details are available on our website: \url{https://sites.google.com/view/ata-hrl}.
\end{abstract}

\section{Introduction}
With the increasing demand for addressing complex, large-scale challenges such as disaster response, search and rescue, and environmental monitoring, multi-human multi-robot (MH-MR) teams have gained significant attention \cite{dahiya2023survey}. By leveraging the diverse skills and expertise of both humans and robots working in tandem, MH-MR teams can enhance team complementarity, productivity, and flexibility \cite{wang2024AeHRL}.

However, the expanded scale and complication of MH-MR teams introduce new challenges in task allocation. One key challenge arises from the intrinsic heterogeneity within MH-MR teams, where each team member brings distinct characteristics and capabilities \cite{dahiya2023survey}. Humans may vary in cognitive abilities and skill levels \cite{humann2018modeling}, while robots can differ in mobility, autonomy, and sensory capabilities \cite{fu2022robust}. At the same time, tasks themselves have unique requirements in terms of complexity, location, and duration \cite{nunes2017taxonomy}. To effectively harness the strengths of each team member while accounting for their limitations, sophisticated task allocation strategies that go beyond simple matching of tasks to team members are necessary.

\begin{figure}[t]
\centering
\includegraphics[width=1\columnwidth]{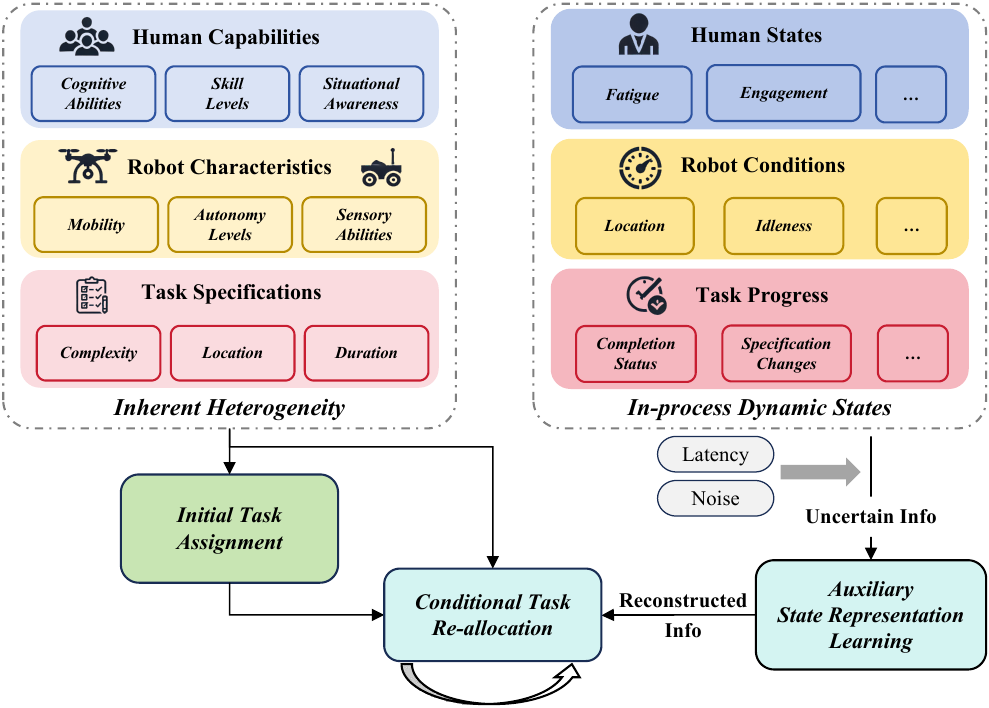}
\vspace{-17pt}
\caption{Conceptual illustration of our adaptive task allocation method, named ATA-HRL, in MH-MR teams. Unlike previous one-sided approaches, we consider both inherent heterogeneity and in-process dynamic states of the team and its assigned tasks, hierarchically combining initial task allocation and conditional task reallocation. To handle state information uncertainty, we also introduce an auxiliary state learning task to contextually reconstruct incomplete or noisy state information.}
\vspace{-20pt}
\label{fig:concept}
\end{figure}

Compounding this challenge is the highly dynamic nature of MH-MR team operations \cite{mina2020adaptive,jo2024mocas}. During task execution, human states, such as fatigue and engagement, can fluctuate over time, and robot operational conditions are also subject to change \cite{jo2024cognitive}. Concurrently, task specifications initially estimated may evolve as the operation progresses, introducing additional dynamics \cite{dias2006market}. More crucially, in real-world scenarios, information about these dynamic states is often uncertain, influenced by inaccuracies in perception and communication latency \cite{chakraa2023optimization}. For instance, human fatigue may not be measured with perfect accuracy, whether assessed subjectively \cite{barua2020towards} or objectively \cite{wang2024husformer}, and updates regarding robot status and task attributes can be delayed \cite{otte2020auctions}. This necessitates adaptive task allocation methods that can respond to dynamic team states and changing conditions of tasks under uncertain information.

Existing literature has typically addressed these challenges in isolation, rarely considering them simultaneously. Studies focused on team heterogeneity have primarily developed task assignment strategies that align agent capabilities with task requirements \cite{zhang2016co,wang2023initial,wang2024AeHRL,humann2023modeling,fu2022robust,ravichandar2020strata}. However, these approaches often overlook how team states evolve during task execution, limiting their applicability in real-world dynamic environments. On the other hand, research on dynamic states primarily targets real-time task allocation adjustments based on changes in team conditions \cite{jo2023affective,mina2020adaptive,wu2022task,10309392,chatzikonstantinou2020integrated}. While these methods can adapt to fluctuations such as human fatigue or robot idleness, they frequently fail to account for the inherent variability in team composition, potentially resulting in prolonged suboptimal performance.

Moreover, existing approaches to handling information uncertainty are insufficient: they either incorporate uncertainty into their optimization process, such as using partially observable Markov decision process (POMDP) to account for state uncertainty in reinforcement learning (RL) \cite{wu2022task}, or treat it as an additional cost to minimize in model-based approaches \cite{li2005robust,elgibreen2019dynamic}. Alternatively, some methods \cite{fu2022robust,rudolph2021desperate} rely on ensuring capability redundancy for each task to improve robustness to uncertainty, which is often impractical in resource-constrained environments.

\begin{figure*}[!t]
\centering
\includegraphics[width=\linewidth]{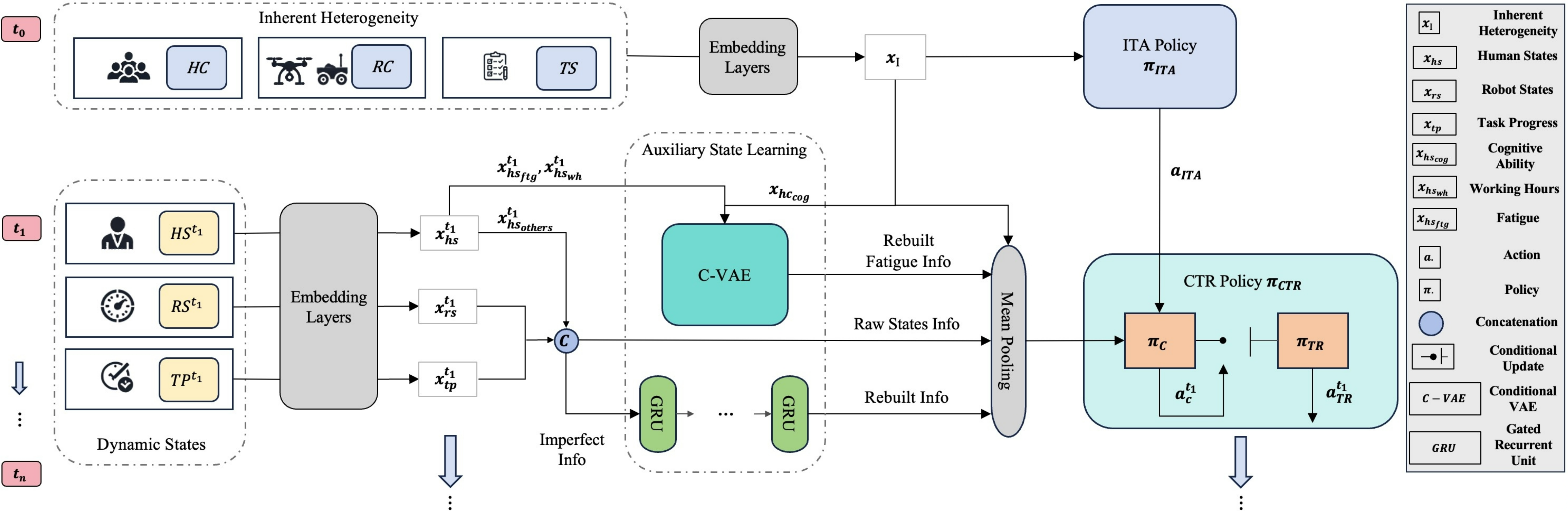}
\vspace{-18pt}
\caption{Illustration of the proposed ATA-HRL framework. The main HRL hierarchy consists of two levels: the first, at time step $0$, determines the optimal ITA by considering inherent team heterogeneity; the second, at each subsequent time step $1-n$ during operation, decides whether to reallocate tasks and how to allocate them, considering additional dynamic operational changes. The optional reallocation decision is represented by a switch icon. An auxiliary state learning module is integrated into the second layer to address state information uncertainty, enhancing decision-making during reallocation.}
\vspace{-15pt}
\label{fig:framework}
\end{figure*}

To address these limitations, as shown in Fig. \ref{fig:concept}, we propose a novel adaptive task allocation method for MH-MR teams that hierarchically considers both team heterogeneity and dynamic in-operation states. Specifically, as illustrated in Fig. \ref{fig:framework}, we design a hierarchical RL framework (named ATA-HRL) with two levels: one for initial task assignment (ITA), aligning team capabilities with task requirements before execution, and another for conditional task reallocation (CTR), which dynamically adjusts to evolving team states. Hierarchical RL (HRL) allows us to break down the complex task allocation problem into manageable sub-tasks, making it more effective for dynamic real-world scenarios. Furthermore, we introduce an auxiliary state representation learning task to manage information uncertainty by contextually reconstructing inaccurate or delayed state observations. Our key contributions are summarized as:
\begin{itemize}[leftmargin=*]
    \item We introduce ATA-HRL to hierarchically account for both inherent heterogeneity and dynamic operational states in MH-MR teams, optimized through a HRL structure.
    \item To handle information uncertainty, we propose an auxiliary state representation learning task that reconstructs uncertain information, enhancing policy learning.
    \item We present a benchmark simulation environment that replicates both the heterogeneity and dynamics of MH-MR teams based on our previous works \cite{wang2023initial,wang2024AeHRL}, and conduct a comprehensive case study to demonstrate the advantages of our ATA-HRL over state-of-the-art approaches.
\end{itemize}

\section{Background}
\label{RW}
\noindent \textbf{HRL} offers an efficient approach to breaking down complex decision-making problems into smaller, manageable sub-tasks \cite{pateria2021hierarchical}. It employs a hierarchy of policies, each addressing a specific facet of the decision-making process. These policies are executed hierarchically, governed by predefined initiation and termination conditions, and trained jointly or separately using standard RL. Our approach follows the HRL paradigm with a two-level policy hierarchy, specifically utilizing the \textit{options} framework \cite{sutton1999between}.

\noindent \textbf{Auxiliary Tasks in RL} have been explored to enhance learning efficiency and robustness by incorporating these supplementary learning targets \cite{jaderberg2016auxiliary}, such as state representation learning \cite{hansen2021generalization}, predicting termination conditions \cite{kartal2019terminal}, and modeling state-action dynamics \cite{zheng2024texttt}. In this work, we construct an auxiliary state representation learning task to address the uncertainty in dynamic operational states of humans, robots, and tasks caused by unavoidable noise and latency in real-world scenarios. This integration enables the reconstruction of contextual information representations that are more robust to noise and incomplete observations. 


\section{Methodology}
\subsection{Problem Formulation}
\label{PF}
Building upon \cite{wang2024AeHRL}, we formulate the task allocation problem in MH-MR teams as a hierarchical Markov decision process (HMDP) that incorporates both ITA and CTR. The HMDP is formally defined as a tuple: $<\mathcal{M}, \mathcal{J}, \mathcal{I}, \mathcal{S}, \mathcal{A}, \mathcal{T}, \mathcal{R}>$, where:
\begin{itemize}[leftmargin=*]
    \item $\mathcal{M} := \{hm_1, \dots, hm_i, rm_1, \dots, rm_j\}$ denotes the heterogeneous members within the MH-MR team, including $i$ human operators and $j$ robots.
    \item $\mathcal{J} := \{ta_1, \dots, ta_k\}$ represents the job that the MH-MR team needs to accomplish, which is subdivided into a finite set of $k$ tasks.
    \item $\mathcal{I} := \{{I}_{hm} \times {I}_{rm} \times {I}_{ta}\}$ denotes the inherent heterogeneity information of the MH-MR team and its designated tasks:
    \begin{itemize}[leftmargin=*]
        \item ${I}_{hm} := \{ \{hc_1^1, \ldots, hc_p^1\} \times \ldots \times \{hc_{1}^i, \ldots, hc_{p}^i\} \}$, representing the diverse individual capabilities of the $i$ human members across $p$ domains, such as skill levels and cognitive abilities in our case study.
        \item ${I}_{rm} := \{ \{rc_1^1, \ldots, rc_q^1\} \times \ldots \times \{rc_{1}^j, \ldots, rc_{q}^j\} \}$, representing the individual characteristics of the $j$ robot members within $q$ contexts, such as are sensory abilities and mobility in our case study.
        \item ${I}_{ta} := \{ \{ts_1^1, \ldots, ts_m^1\} \times \ldots \times \{ts_{1}^k, \ldots, ts_{m}^k\} \}$, representing the individual specifications of the $k$ tasks across $m$ contexts, such as complexity and spatial attributes in our case study.
    \end{itemize}
    
In line with \cite{humann2018modeling,humann2023modeling,wang2023initial}, we assume that ${I}_{hm}$ and ${I}_{rm}$ remain unchanged during operation and can be assessed well in prior, but ${I}_{ta}$ may change due to initial inaccuracies in estimation or unforeseen task-related dynamics. These dynamics are modeled within ${S}_{ta}$ as described below.
   \item $\mathcal{S} := \{{S}_{hm} \times {S}_{rm} \times {S}_{ta}\}$ is the observation of dynamic states of the team and tasks during operations:
    \begin{itemize}[leftmargin=*]
        \item ${S}_{hm} := \{ \{hs_1^1, \ldots, hs_u^1\} \times \ldots \times \{hs_1^i, \ldots, hs_u^i\} \}$, representing the collective states of the $i$ human members across $u$ domains, which in our case study are working time, fatigue, idleness and situational awareness.
        \item ${S}_{rm} := \{ \{rs_1^1, \ldots, rs_v^1\} \times \ldots \times \{rs_1^j, \ldots, rs_v^j\} \}$, representing the status of the $j$ robot members across $v$ contexts, which in our case study are positions, idleness, and working conditions.
        \item ${S}_{ta} := \{ \{tp_1^1, \ldots, tp_w^1\} \times \ldots \times \{tp_1^k, \ldots, tp_w^k\} \}$, representing the progress of the $k$ tasks across $w$ contexts, which in our case study are task completion status and specification changes.
    \end{itemize}
    
Unlike $\mathcal{I}$, we assume that $\mathcal{S}$ changes at each step during operations, with uncertainty arising from two key aspects. First, noise affects measurements such as human fatigue in $\mathcal{S}_{hm}$, leading to inaccurate observations. Second, latency occurs in other factors within $\mathcal{S}_{hm}$, $\mathcal{S}_{rm}$, and $\mathcal{S}_{ta}$, particularly when team members conducting tasks in the field, where long-range information transmission is required, increasing the delay in state updates.

\item $\mathcal{A} := \{A_{\text{ITA}} \times A_{\text{CTR}}\}$ represents the hierarchical task assignment decisions, consisting of two components:
\begin{itemize}
    \item $A_{\text{ITA}}$ = $a_{\text{ITA}}^{t_{0}}$ is an ITA action determined by an ITA policy $\pi_{\text{ITA}}: \mathcal{I} \mapsto a_{\text{ITA}}$, which maps the team and task heterogeneity observation to an ITA decision only at step $0$, i.e., before task execution begins.
    \item $A_{\text{CTR}} := \{A_{\text{C}} \times A_{\text{TR}}\}$ denotes a series of joint CTR actions at every step during operations, comprising two sub-components. $A_{\text{C}} = \sum_{t=1}^{T} a_{\text{C}}^{t}$ refers to the series of decisions on whether reallocation is necessary, determined by the reallocation condition policy $\pi_{\text{C}}: \mathcal{I} \times \mathcal{S} \times a^{t_{\text{Prev}}}_{\text{TR}} \mapsto a_{\text{C}}$. And $A_{\text{TR}} = \sum_{t=1}^{T} a_{\text{TR}}^{t}$ refers to the series of task reallocation decisions at each time step, determined by the task reallocation policy $\pi_{\text{TR}}: \mathcal{I} \times \mathcal{S} \times a^{t_{\text{Prev}}}_{\text{TR}} \mapsto a_{\text{TR}}$, which decides how tasks should be reassigned if reallocation is triggered. Here, $a^{t_{\text{Prev}}}_{\text{TR}}$ is the previous task reallocation decision, which is equivalent to $a_{\text{ITA}}^{t_{0}}$ at $t_1$.
\end{itemize}
These sub-policies are structured hierarchically for both training and execution based on the classic \textit{options} framework in HRL \cite{sutton1999between}, where the starting and termination flags of each policy are manually defined.
    \item \(\mathcal{T} := \mathcal{P}(\mathcal{S}^{\prime} \mid \mathcal{S})\) denotes the unknown probabilistic state transition function, representing the likelihood of transitioning from the current state \( \mathcal{S} \) to a new state \( \mathcal{S}^{\prime} \).
    \item $\mathcal{R} = R_{\text{ITA}} \times R_{\text{C}} \times R_{\text{TR}}$ is the joint reward function, providing individual rewards for each sub-policy in the hierarchy. See Sec. \ref{reward} for details.
\end{itemize}
The objective is to find a set of optimal policies $\pi_{\text{ITA}}$, $\pi_C$, and $\pi_{TR}$ that jointly maximize the expected sum of rewards, $\mathbb{E}\left[\sum_{t} \mathcal{R}\right]$. This formulation explicitly models the inherent heterogeneity and in-operation dynamic states of the MH-MR team and its assigned tasks. We aim to address this HMDP with an HRL-based framework, ATA-HRL, whose details are introduced below.

\subsection{Initial Task Assignment}
The initial level in ATA-HRL is to initialize task assignments that consider the heterogeneity in the team and its designated tasks. To this end, we pass the heterogeneity information \( \mathcal{I} = \{ I_{hc} \times I_{rc} \times I_{ts} \} \) through temporal embedding layers \cite{wang2024AeHRL}, \( f_{emb} \), where each component is transformed into a common dimensional space and added with positional embeddings $E_{h,r,t}$ to introduce awareness of the order of humans, robots, and tasks as:
\begin{equation}
    x_{I_{hc}, I_{rc}, I_{ts}} = f_{emb}^{hc, rc, ts}(I_{{hc, rc, ts}}) + E_{h,r,t}
\end{equation}

The embedded representations, \( x_{I} \), are then used as input to an ITA policy $\pi_{\text{ITA}}$, which is implemented as an attention-based RL network as in \cite{wang2023initial}, generating an ITA action $\pi_{\text{ITA}}$. For more details, we refer readers to the original paper. Note that this initialization only occurs once before task execution.

\subsection{Conditional Task Reallocation with State Reconstruction}
The next level in ATA-HRL is to adaptively reallocate task distributions, which is the ITA at time step $1$, during task execution to the end. At each step, we use the same embedding layers to generate representations of dynamic states:
\begin{equation}
    x_{S_{hs}, S_{rs}, S_{tp}} = f_{emb}^{hs, rs, tp}(S_{{hs, rs, tp}}) + E_{h,r,t}
\end{equation}

\begin{figure}[t]
\centering
\includegraphics[width=1\columnwidth]{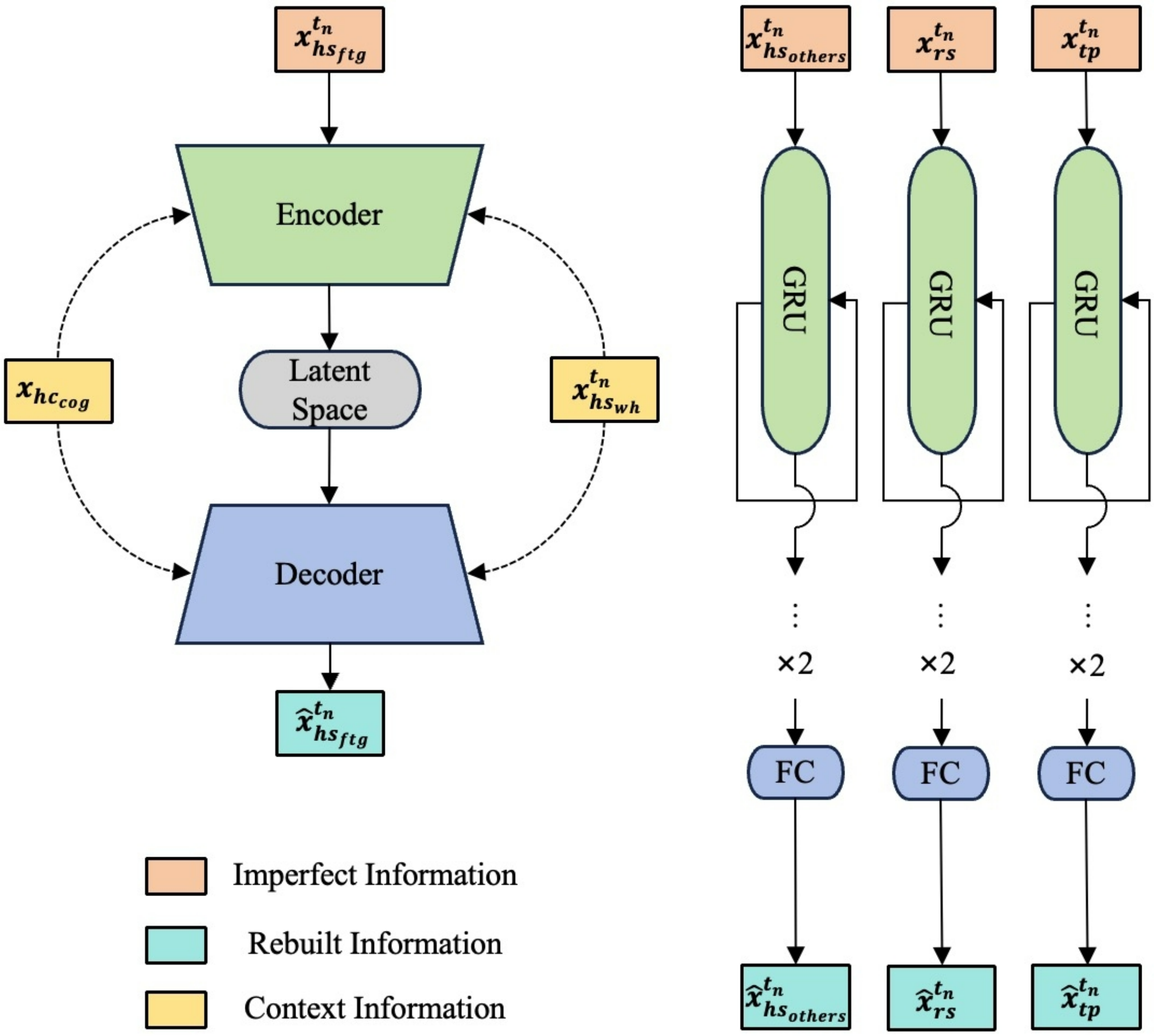}
\vspace{-17pt}
\caption{Detailed structure of the Conditional VAE (left) and GRU-based (right) state reconstruction framework.}
\vspace{-20pt}
\label{fig:rebuilt}
\end{figure}

However, as mentioned earlier, the state information is subject to uncertainty, affected by unavoidable noise in human fatigue perception and latency in the states of agents working in the field. To address this, we add extra state representation reconstruction modules before the CTR policies as an auxiliary task to learn. Specifically, as shown in Fig. \ref{fig:rebuilt} left, we utilize a conditional Variational Autoencoder (cVAE)\cite{yu2022structure} to infer contextual representations of human fatigue observations, which are learned conditioned on other related factors, such as working hours and cognitive abilities, which can be observed more reliably:
\begin{equation}
    \hat{x}_{hs_{{ftg}}} = D(E(x_{hs_{wh}}, x_{hc_{cog}}; \theta_E), x_{hs_{wh}}, x_{hc_{cog}}; \theta_D)
\end{equation}
where, \(E\) and \(D\) refer to the encoder and decoder in the cVAE structure. \(\hat{x}_{hs_{{ftg}}}\) is the reconstructed representation of human fatigue, and \(\theta_E\) and \(\theta_D\) denote the learnable parameters of the encoder and decoder.

Our rationale is that these condition factors are correlated with actual fatigue \cite{enoka2016translating}, allowing the model to learn a more meaningful representation of fatigue. Importantly, the goal is not to reconstruct the exact fatigue level, but to generate a more efficient contextual representation within the current MDP dynamics.

Furthermore, as shown in Fig. \ref{fig:rebuilt} (right), to handle latency, we follow the approach in \cite{hao2021hierarchical} and design a model that stacks three GRU units with a feedforward layer for reconstructing the missing or delayed states as follows:
\begin{equation}
    \hat{x}_{.} = \text{FeedForward}(\text{GRU}_3(\text{GRU}_2(\text{GRU}_1(x_{.}; \theta_{\text{GRU}}))))
\end{equation}
where \(x_{.}\), represents the input robot, human, or task state, and \(\theta_{\text{GRU}}\) denotes the learnable parameters shared across the three GRU units and the feedforward layer. 

This stacked architecture helps process time-related sequential information more effectively through the GRU layers, which are then refined by the feedforward layer for enhanced accuracy in state reconstruction.

After obtaining all reconstructed state representations, we concatenate them with the raw state representations and pass them through a mean pooling layer for dimensionality reduction. The process is represented as:
\begin{equation}
    \hat{x}_{\text{final}} = \text{MeanPool}([\hat{x}_{.} \| x_{.}])
\end{equation}
where \(\hat{x}_{.}\) represents the reconstructed state, \(x_{.}\) represents the raw state, \(\|\) denotes concatenation, and \(\text{MeanPool}\) is the mean pooling operation applied to the concatenated representations.

The output state representation, along with the inherent information and task allocation from the previous step, forms the current state input of \(\pi_{\text{CTR}}\), which consists of two sub-policies: \(\pi_C\) and \(\pi_{TR}\), both implemented as GRU networks. The sub-policy \(\pi_C\) decides whether task reallocation is necessary, and the sub-policy \(\pi_{TR}\) determines how tasks should be reallocated among the team members to optimize performance under the current state conditions. If \(\pi_C\) decides that reallocation is unnecessary, the execution of \(\pi_{TR}\) is skipped. This two-step process ensures that tasks are not reallocated too frequently, maintaining smooth operations.

\subsection{Reward Shaping and Model Training}
\label{reward}
To effectively train the HRL network, each sub-policy must have a well-defined reward function. We first assume a time-step-based reward \(r^{\text{perf}}_t\), which reflects the task performance at time step \(t\) during execution, e.g., classification accuracy in our environmental monitoring case study.

We shape the reward function \(R_{\text{ITA}}\) for the ITA policy to maximize overall task performance while penalizing costly reallocations during task execution:
\begin{equation}
R_{\text{ITA}} =  \sum_{t=0}^{T} r^{\text{perf}}_t - 0.2 \cdot \sum_{t=1}^{T} \| a_{\text{TR}}^{(t)} - a_{\text{ITA}} \|_F
\end{equation}
where \(\sum_{t=0}^{T} r^{\text{perf}}_t\) represents the total task performance from the initial to the final step, and \(\| a_{\text{TR}}^{(t)} - a_{\text{ITA}} \|_F\) is the Frobenius norm \cite{bottcher2008frobenius} of the difference between the TR and ITA at each time step when TR is triggered.

The reward for the condition policy \(R_{\text{C}}\) is defined as:
\begin{equation}
R_{\text{C}} =  \sum_{t=0}^{T} r^{\text{perf}}_t + \sum_{t_{\text{prev}}}^{t_{\text{cur}}} \Delta r^{\text{perf}} - 0.4 \cdot \frac{F_{\text{realloc}}}{T}
\end{equation}
where \( \Delta r^{\text{perf}} = r^{\text{perf}}_{t_{\text{cur}}} - r^{\text{perf}}{t_{\text{prev}}} \) measures the performance improvement between reallocations, calculated as the difference in performance after the current reallocation and the previous one; \( \frac{F_{\text{realloc}}}{T} \) is the normalized frequency of reallocations, penalizing excessive reallocations.

The reward for the task reallocation policy \(R_{\text{TR}}\) is defined as:
\begin{equation}
R_{\text{TR}} =  \sum_{t=0}^{T} r^{\text{perf}}_t + \sum_{t_{\text{prev}}}^{t_{\text{cur}}} \Delta r^{\text{perf}} - 0.2 \cdot \frac{\sum_{t_{\text{prev}}}^{t_{\text{cur}}} \Delta a_{\text{realloc}}}{T}
\end{equation}
where \( \frac{\sum_{t_{\text{prev}}}^{t_{\text{cur}}} \Delta a_{\text{realloc}}}{T} \) represents the normalized sum of task reallocation changes, where \( \Delta a_{\text{realloc}} = \| a_{\text{TR}}^{(t)} - a_{\text{TR}}^{(t-1)} \|_F \) is the Frobenius norm of the change in task allocation between consecutive time steps, penalizing dramatic reallocations.

These two rewards enable the entire CTR policy to balance performance improvement with reallocation efficiency. Furthermore, to optimize the cVAE and GRU-based networks in the auxiliary state representation learning module, we designed auxiliary losses. For the cVAE, we used the conventional reconstruction loss and KL divergence loss:
\begin{equation}
\mathcal{L}_{\text{cVAE}} = \| x_{hs_{{ftg}}} - \hat{x}_{hs_{{ftg}}} \|^2_2 + 0.1 \cdot \mathcal{L}_{\text{KL}}
\end{equation}
where the first term represents the reconstruction loss, measuring the difference between the fatigue input \(x_{hs_{{ftg}}}\) and its reconstructed output \(\hat{x}_{hs_{{ftg}}}\), and $\mathcal{L}_{\text{KL}}$ denotes the common KL divergence loss in VAE structures \cite{kingma2013auto} to regularize the latent space.

While this self-supervised loss does not directly optimize the state representation for task performance, it works jointly with the RL optimization process to capture meaningful latent state representations as seen in \cite{allshire2021laser}. To train the GRU-based network, since we have access to the ground-truth data for latency (as the actual state will eventually be conveyed or logged back), we treat this as a supervised learning task before being used in the main RL loop via a mean squared error (MSE) loss, following \cite{hao2021hierarchical,zheng2024texttt}.


\section{Case Study and Experiments}
\label{case}

\subsection{Task Scenario} \label{caseset} To assess the effectiveness of our ATA-HRL framework, we developed a case study scenario based on \cite{wang2024AeHRL}. The scenario involves a large-scale environmental surveillance task focused on monitoring pollution hazards leaking from warehouses and factories. The mission begins when a satellite system detects various POIs, representing pollution on the ground or in the air. The hazard classification is divided into three difficulty levels: low, medium, and high. The MH-MR team needs to: (1) Navigating to each POI to capture images, which can be done autonomously by robots or through collaborative control, where human operators provide essential navigation and imaging guidance; and (2) Analyzing the captured images to determine whether the POI represents an actual hazard, a task performed either by human experts or using onboard object detection algorithms.

Correct identifications award points; 15 for low difficulty, 25 for medium, and 35 for high. Incorrect identifications result in deductions of the same amounts. The success rates are dependent on human and robot performance models, which are discussed in subsequent sections. We define the reward \(r^{\text{perf}}_t\) in Sec. \ref{reward} as the total points gained by the team at each time step.

\subsection{Simulation Environment} 
\begin{figure}[t]
\centering
\includegraphics[width=0.95\columnwidth]{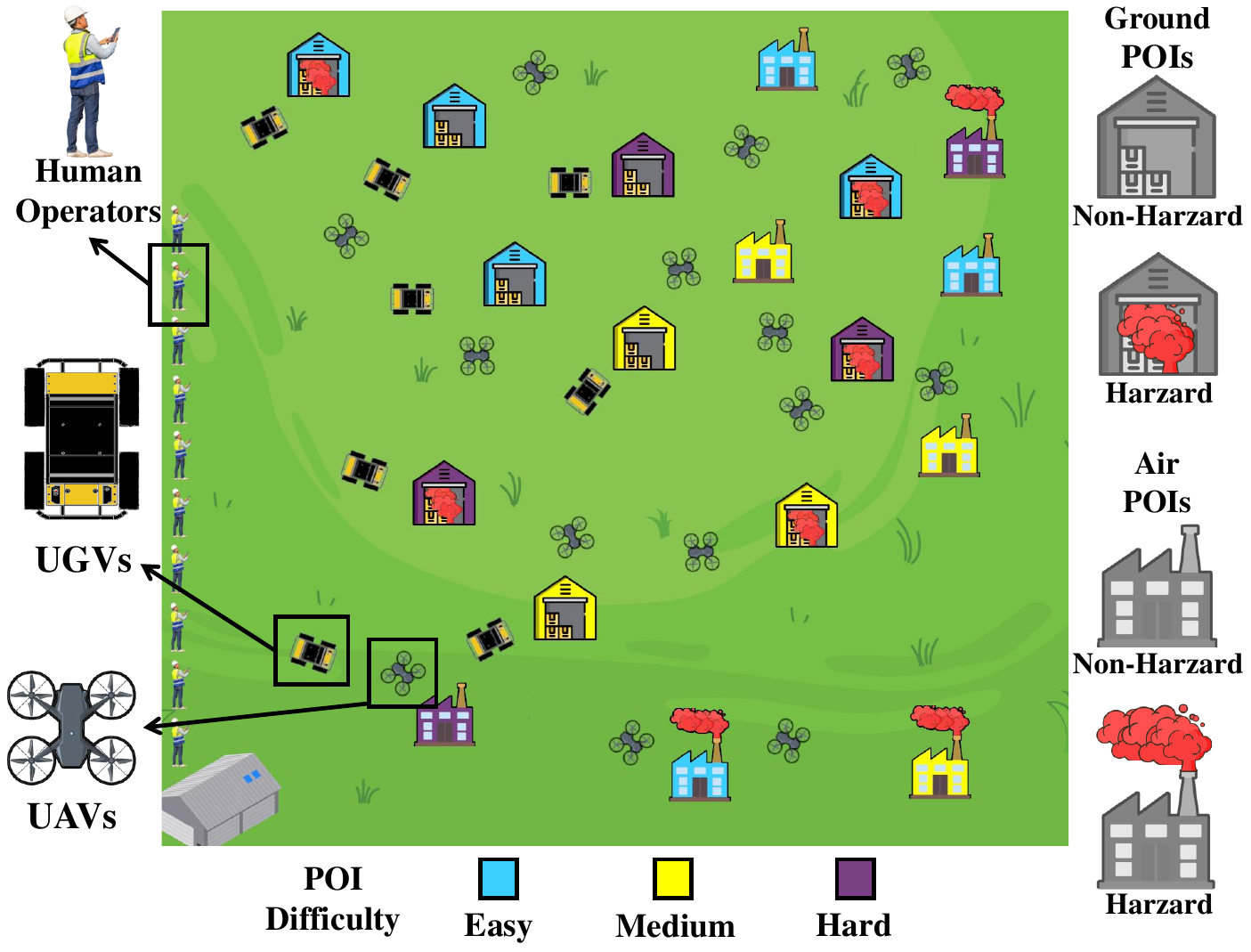}
\vspace{-10pt}
\caption{Visual illustration of the simulation environment, zoomed for visibility. Each POI is distinguished by color to indicate the complexity level for hazard evaluation. Additionally, the type of pollution at each POI is determined by the building type, with warehouses representing ground POIs and factories representing air POIs.}
\label{fig:sim}
\vspace{-10pt}
\end{figure}

We adopted a benchmark simulation environment for ITA in MH-MR teams \cite{wang2023initial,wang2024AeHRL,humann2023modeling} to add additional factors to model operational dynamic states and uncertainty. We refer readers to \cite{wang2024AeHRL} for details of the team heterogeneity settings. As shown in Fig. \ref{fig:sim}, the environment covers a 2~$km \times$ 2~$km$ area designed for environmental surveillance, featuring multiple POIs with varying hazard types, air or ground POIs, and classification difficulties.

\subsubsection{Robot Model} We consider two types of robots, UAVs and UGVs, with different mobility and sensor capabilities. In autonomous mode, UAVs are configured with faster default speed parameters. The default image quality of the robots varies depending on the type of POI being observed, with UAVs being more effective for air pollution monitoring and UGVs better suited for ground pollution. These parameters are further influenced by the operator's level of expertise in collaborative operation mode, where higher proficiency improves both speed and image quality. Furthermore, the probability of successful onboard hazard classification by the robot is influenced by both the captured image quality and the inherent difficulty of the hazard. Detailed specifications for speed and robot classification success probability \(\mathcal{P}_{rc}\) can be found in \cite{wang2024AeHRL}, while Table \ref{tab:image_quality} provides the new settings of image quality in our work. 

\subsubsection{Human Model} We model human operators as sequential event handlers \cite{watson2017informing}, with their performance in hazard classification influenced by factors such as fatigue, workload, and task complexity, which are further moderated by their cognitive abilities and skill levels. The probability of accurate hazard classification by a human is determined nonlinearly \cite{humann2018modeling} based on these elements as:
\begin{equation}
\small
\mathcal{P}_{hc} =\frac{1}{2}+\eta (F_f F_w) \cdot \lambda (F_d)
\label{HP}
\end{equation}
where $\eta, \lambda \in (0,\frac{\sqrt{2}}{2})$, are the adjustment weights corresponding to individual cognitive capacity and operational expertise, respectively. $F_f \in (0,1]$ and $F_w$, $F_d$ $\in (0,1)$ denote the influences of fatigue, workload, and decision-making challenges, respectively. This formulation ensures that the minimum probability of successful classification approaches 0.5, reflecting the random binary guessing \cite{humann2018modeling,humann2023modeling}. We follow the same approach to calculate $F_f$, $F_w$ and $F_d$ as stated in\cite{wang2024AeHRL}, as well as the value settings of $\eta$ and $\lambda$. 

\begin{table}[t]
\centering
\caption{Quality of Captured Image for Ground and Air Pollution: image quality for both robots in Auto and H-LS, H-MS, and H-HS conditions.}
\vspace{-5pt}
\resizebox{\linewidth}{!}{
\begin{tabular}{c ||ccc |ccc}
\toprule
 \multirow{2}{*}{\textit{Robot Type}} & \multicolumn{3}{c|}{\textit{Ground Pollution}} & \multicolumn{3}{c}{\textit{Air Pollution}} \\
\cmidrule(lr){2-4} \cmidrule(lr){5-7}
 & H-LS & Auto / H-MS & H-HS & H-LS & Auto / H-MS & H-HS \\
\midrule
UAV & Low & Medium & Upper-Medium & Medium & Upper-Medium & High \\
UGV & Medium & Upper-Medium & High & Low & Medium & Upper-Medium \\
\bottomrule
\end{tabular}
}
\label{tab:image_quality}
\vspace{-15pt}
\end{table}

\subsubsection{Modeling Information Uncertainty}\label{uncertain} In contrast to \cite{wang2024AeHRL}, we introduce various forms of uncertainty and dynamic elements to simulate the unpredictability and challenges of real-world operations. First, to model the uncertainty in estimating task attributes during the ITA phase, specifically in satellite-based estimation of POI types and classification difficulties, we introduce random changes in the estimated POI attributes as the operation progresses. These changes are modeled as random events. Another type of random event involves robot failures, where robots may randomly stop functioning, simulating unexpected equipment malfunctions. Both types of changes are incorporated into the dynamic states $\mathcal{S}$ at the next timestep after they occur.

Moreover, random latency is introduced in ${S}_{rm,ta}$ for parameters such as current location, speed, and working conditions. At random intervals, based on a predefined probability, these parameters are replaced with values from the previous timestep to simulate unexpected delays or performance disruptions. This latency does not apply to human states, as the humans in our case work onsite, minimizing delays in information regarding working hours and utilization. However, to simulate imperfect estimations of human fatigue, Gaussian noise is added to the human fatigue calculation at each timestep.

\subsection{Experiments and Results}
\begin{figure*}[t]
\centering
\includegraphics[width=0.9\linewidth]{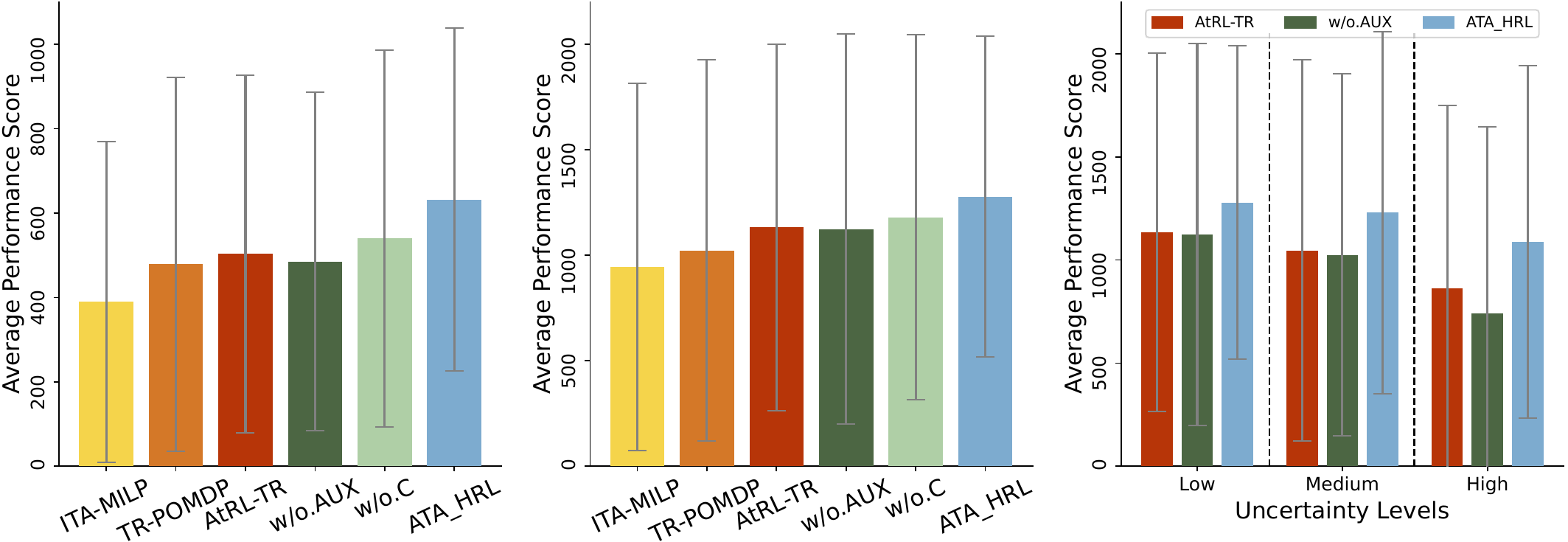}
\vspace{-10pt}
\caption{Comparison of ATA-HRL with baselines and ablation models in the setting of with 6 humans, 8 robots, and 60 POIs (left) and setting of with  12 humans, 16 robots, and 130 POIs (middle), and ablation study results (right).}
\vspace{-20pt}
\label{fig:results}
\end{figure*}

\subsubsection{Baselines and Ablation Models}
We compared our proposed ATA-HRL method against three baselines: i) ITA-MILP: This baseline uses a state-of-the-art (SOTA) mixed-integer linear programming (MILP) approach \cite{fu2022robust} for ITA, that considers information uncertainty by ensuring ability redundancy into the ITA for each task. Originally designed for multi-robot systems, we adapt this method to our setting by treating humans as another type of robot; ii) TR-POMDP: This baseline represents a SOTA in-process task reallocation (TR) method \cite{wu2022task}, utilizing decentralized DQN and handling uncertainty with a POMDP formulation; and iii) AtRL-TR: We build this by combining the AtRL \cite{wang2023initial} for ITA with the TR-POMDP \cite{wu2022task} for TR, providing a baseline lacking in the literature that addresses both ITA and TR. 

To further assess the contribution of each component within our method, we introduced two ablation models: i) w/o.AUX: it removes the auxiliary task representation learning module from ATA-HRL; and ii) w/o.C: it removes the sub-policy for reallocation timing determination in ATA-HRL, forcing task reallocation at every timestep.

To evaluate the robustness of our method under uncertain conditions, we consider three levels of uncertainty: low, medium, and high. These levels are modeled by adding different Gaussian noise levels \((\sigma^2 = 0.05, 0.1, 0.2)\) to human fatigue, and by applying different probabilities of random events \((0.1, 0.2, 0.4)\) as mentioned in Sec.~\ref{uncertain}.

\subsubsection{Evaluation}
We set two MH-MR team settings: (a) 6 humans, 8 robots, and 60 POIs; (b) 12 humans, 16 robots, and 130 POIs, testing each method across 500 unseen scenarios, where attributes of humans, robots, and POIs were randomly generated. We reported the average total team performance scores \(\left( \sum_{t=0}^{T} r^{\text{perf}}_t \right)\) as evaluation metrics.

\subsubsection{Results and Analysis}
\label{results}
The performance of our proposed ATA-HRL compared to baselines across settings (a) and (b) is depicted in Fig. \ref{fig:results} (left and middle). We observe that ATA-HRL consistently outperforms other methods in terms of average performance scores, leading to better overall team performance across medium and large MH-MR teams. 

Specifically, the ITA-MILP method, which focuses only on ITA under team heterogeneity, performs poorly. We attribute this to the fact that, while it leverages ability-redundant ITA solutions to cover dynamic operational states, it does not explicitly account for task reallocation. This becomes more critical in our case, where the dynamics of MH-MR team states and changes in task attributes are heavily modeled. Similarly, the TR-POMDP method also exhibits performance drops, likely due to its reliance on a randomly generated ITA, neglecting the importance of ITA. In contrast to these one-sided approaches, ATA-HRL hierarchically executes ITA followed by task reallocation to handle the evolving states of MH-MR teams. These observations highlight the benefits of simultaneously addressing both inherent heterogeneity and dynamic operational states.

Moreover, while AtRL+TR outperforms the one-sided baselines, it still lags behind ATA-HRL. This is because, although it considers both ITA and reallocation, it does not explicitly handle information uncertainty. In contrast, our ATA-HRL leverages an auxiliary task to reconstruct state representations, leading to more robust performance. This advantage is further evidenced by the performance drop in the ablation model w/o.AUX. Furthermore, as uncertainty increases, ATA-HRL shows stronger performance improvements, as shown in Fig. \ref{fig:results} (right). This highlights the benefits of the auxiliary state represent model proposed. In addition, compared to the ablation model w/o.C, our ATA-HRL shows clear performance improvements. This is because we introduce an additional sub-policy for determining whether reallocation is necessary. This enables more informed decision-making, ensuring that tasks are not reallocated unnecessarily, thus optimizing team efficiency under dynamic conditions.

\section{Conclusion}
This paper presents ATA-HRL, a novel adaptive task allocation with HRL for MH-MR teams, which effectively integrates initial task assignment and conditional reallocation while addressing state uncertainty. Experimental results demonstrate the robustness and adaptability of ATA-HRL to inherent heterogeneity and dynamic information uncertainty.

\typeout{}
\bibliography{main}

\begin{thebibliography}{10}
\providecommand{\url}[1]{#1}
\csname url@samestyle\endcsname
\providecommand{\newblock}{\relax}
\providecommand{\bibinfo}[2]{#2}
\providecommand{\BIBentrySTDinterwordspacing}{\spaceskip=0pt\relax}
\providecommand{\BIBentryALTinterwordstretchfactor}{4}
\providecommand{\BIBentryALTinterwordspacing}{\spaceskip=\fontdimen2\font plus
\BIBentryALTinterwordstretchfactor\fontdimen3\font minus \fontdimen4\font\relax}
\providecommand{\BIBforeignlanguage}[2]{{%
\expandafter\ifx\csname l@#1\endcsname\relax
\typeout{** WARNING: IEEEtran.bst: No hyphenation pattern has been}%
\typeout{** loaded for the language `#1'. Using the pattern for}%
\typeout{** the default language instead.}%
\else
\language=\csname l@#1\endcsname
\fi
#2}}
\providecommand{\BIBdecl}{\relax}
\BIBdecl

\bibitem{dahiya2023survey}
A.~Dahiya, A.~M. Aroyo, K.~Dautenhahn, and S.~L. Smith, ``A survey of multi-agent human--robot interaction systems,'' \emph{Robotics and Autonomous Systems}, vol. 161, p. 104335, 2023.

\bibitem{wang2024AeHRL}
R.~Wang, D.~Zhao, A.~Gupte, and B.-C. Min, ``Initial task allocation in multi-human multi-robot teams: An attention-enhanced hierarchical reinforcement learning approach,'' \emph{IEEE Robotics and Automation Letters}, vol.~9, no.~4, pp. 3451--3458, 2024.

\bibitem{humann2018modeling}
J.~Humann and E.~Spero, ``Modeling and simulation of multi-uav, multi-operator surveillance systems,'' in \emph{2018 Annual IEEE International Systems Conference (SysCon)}.\hskip 1em plus 0.5em minus 0.4em\relax IEEE, 2018, pp. 1--8.

\bibitem{fu2022robust}
B.~Fu, W.~Smith, D.~M. Rizzo, M.~Castanier, M.~Ghaffari, and K.~Barton, ``Robust task scheduling for heterogeneous robot teams under capability uncertainty,'' \emph{IEEE Transactions on Robotics}, vol.~39, no.~2, pp. 1087--1105, 2022.

\bibitem{nunes2017taxonomy}
E.~Nunes, M.~Manner, H.~Mitiche, and M.~Gini, ``A taxonomy for task allocation problems with temporal and ordering constraints,'' \emph{Robotics and Autonomous Systems}, vol.~90, pp. 55--70, 2017.

\bibitem{mina2020adaptive}
T.~Mina, S.~S. Kannan, W.~Jo, and B.-C. Min, ``Adaptive workload allocation for multi-human multi-robot teams for independent and homogeneous tasks,'' \emph{IEEE Access}, vol.~8, pp. 152\,697--152\,712, 2020.

\bibitem{jo2024mocas}
W.~Jo, R.~Wang, G.-E. Cha, S.~Sun, R.~K. Senthilkumaran, D.~Foti, and B.-C. Min, ``Mocas: A multimodal dataset for objective cognitive workload assessment on simultaneous tasks,'' \emph{IEEE Transactions on Affective Computing}, 2024.

\bibitem{jo2024cognitive}
W.~Jo, R.~Wang, B.~Yang, D.~Foti, M.~Rastgaar, and B.-C. Min, ``Cognitive load-based affective workload allocation for multihuman multirobot teams,'' \emph{IEEE Transactions on Human-Machine Systems}, 2024.

\bibitem{dias2006market}
M.~B. Dias, R.~Zlot, N.~Kalra, and A.~Stentz, ``Market-based multirobot coordination: A survey and analysis,'' \emph{Proceedings of the IEEE}, vol.~94, no.~7, pp. 1257--1270, 2006.

\bibitem{chakraa2023optimization}
H.~Chakraa, F.~Gu{\'e}rin, E.~Leclercq, and D.~Lefebvre, ``Optimization techniques for multi-robot task allocation problems: Review on the state-of-the-art,'' \emph{Robotics and Autonomous Systems}, p. 104492, 2023.

\bibitem{barua2020towards}
S.~Barua, M.~U. Ahmed, and S.~Begum, ``Towards intelligent data analytics: A case study in driver cognitive load classification,'' \emph{Brain sciences}, vol.~10, no.~8, p. 526, 2020.

\bibitem{wang2024husformer}
R.~Wang, W.~Jo, D.~Zhao, W.~Wang, A.~Gupte, B.~Yang, G.~Chen, and B.-C. Min, ``Husformer: A multi-modal transformer for multi-modal human state recognition,'' \emph{IEEE Transactions on Cognitive and Developmental Systems}, 2024.

\bibitem{otte2020auctions}
M.~Otte, M.~J. Kuhlman, and D.~Sofge, ``Auctions for multi-robot task allocation in communication limited environments,'' \emph{Autonomous Robots}, vol.~44, pp. 547--584, 2020.

\bibitem{zhang2016co}
C.~Zhang and J.~A. Shah, ``Co-optimizating multi-agent placement with task assignment and scheduling.'' in \emph{IJCAI}, 2016, pp. 3308--3314.

\bibitem{wang2023initial}
R.~Wang, D.~Zhao, and B.-C. Min, ``Initial task allocation for multi-human multi-robot teams with attention-based deep reinforcement learning,'' in \emph{2023 IEEE/RSJ International Conference on Intelligent Robots and Systems (IROS)}, 2023.

\bibitem{humann2023modeling}
J.~Humann, T.~Fletcher, and J.~Gerdes, ``Modeling, simulation, and trade-off analysis for multirobot, multioperator surveillance,'' \emph{Systems Engineering}, 2023.

\bibitem{ravichandar2020strata}
H.~Ravichandar, K.~Shaw, and S.~Chernova, ``Strata: unified framework for task assignments in large teams of heterogeneous agents,'' \emph{Autonomous Agents and Multi-Agent Systems}, vol.~34, pp. 1--25, 2020.

\bibitem{jo2023affective}
W.~Jo, R.~Wang, B.~Yang, D.~Foti, M.~Rastgaar, and B.-C. Min, ``Affective workload allocation for multi-human multi-robot teams,'' \emph{arXiv preprint arXiv:2303.10465}, 2023.

\bibitem{wu2022task}
H.~Wu, A.~Ghadami, A.~E. Bayrak, J.~M. Smereka, and B.~I. Epureanu, ``Task allocation with load management in multi-agent teams,'' in \emph{2022 International Conference on Robotics and Automation (ICRA)}.\hskip 1em plus 0.5em minus 0.4em\relax IEEE, 2022, pp. 8823--8830.

\bibitem{10309392}
M.~Lippi, J.~Gallou, J.~Palmieri, A.~Gasparri, and A.~Marino, ``Human-multi-robot task allocation in agricultural settings: a mixed integer linear programming approach,'' in \emph{2023 32nd IEEE International Conference on Robot and Human Interactive Communication (RO-MAN)}, 2023, pp. 1056--1062.

\bibitem{chatzikonstantinou2020integrated}
I.~Chatzikonstantinou, I.~Kostavelis, D.~Giakoumis, and D.~Tzovaras, ``Integrated topological planning and scheduling for orchestrating large human-robot collaborative teams,'' in \emph{Conference on Biomimetic and Biohybrid Systems}.\hskip 1em plus 0.5em minus 0.4em\relax Springer, 2020, pp. 23--35.

\bibitem{li2005robust}
D.~Li and J.~Cruz, ``A robust hierarchical approach to multi-stage task allocation under uncertainty,'' in \emph{Proceedings of the 44th IEEE Conference on Decision and Control}.\hskip 1em plus 0.5em minus 0.4em\relax IEEE, 2005, pp. 3375--3380.

\bibitem{elgibreen2019dynamic}
H.~ElGibreen and K.~Youcef-Toumi, ``Dynamic task allocation in an uncertain environment with heterogeneous multi-agents,'' \emph{Autonomous Robots}, vol.~43, pp. 1639--1664, 2019.

\bibitem{rudolph2021desperate}
M.~Rudolph, S.~Chernova, and H.~Ravichandar, ``Desperate times call for desperate measures: Towards risk-adaptive task allocation,'' in \emph{2021 IEEE/RSJ International Conference on Intelligent Robots and Systems (IROS)}.\hskip 1em plus 0.5em minus 0.4em\relax IEEE, 2021, pp. 2592--2597.

\bibitem{pateria2021hierarchical}
S.~Pateria, B.~Subagdja, A.-h. Tan, and C.~Quek, ``Hierarchical reinforcement learning: A comprehensive survey,'' \emph{ACM Computing Surveys (CSUR)}, vol.~54, no.~5, pp. 1--35, 2021.

\bibitem{sutton1999between}
R.~S. Sutton, D.~Precup, and S.~Singh, ``Between mdps and semi-mdps: A framework for temporal abstraction in reinforcement learning,'' \emph{Artificial intelligence}, vol. 112, no. 1-2, pp. 181--211, 1999.

\bibitem{jaderberg2016auxiliary}
M.~Jaderberg, V.~Mnih, W.~M. Czarnecki, T.~Schaul, J.~Z. Leibo, D.~Silver, and K.~Kavukcuoglu, ``Reinforcement learning with unsupervised auxiliary tasks,'' \emph{arXiv preprint arXiv:1611.05397}, 2016.

\bibitem{hansen2021generalization}
N.~Hansen and X.~Wang, ``Generalization in reinforcement learning by soft data augmentation,'' in \emph{2021 IEEE International Conference on Robotics and Automation (ICRA)}.\hskip 1em plus 0.5em minus 0.4em\relax IEEE, 2021, pp. 13\,611--13\,617.

\bibitem{kartal2019terminal}
B.~Kartal, P.~Hernandez-Leal, and M.~E. Taylor, ``Terminal prediction as an auxiliary task for deep reinforcement learning,'' in \emph{Proceedings of the 15th AAAI Conference on Artificial Intelligence and Interactive Digital Entertainment (AIIDE)}, 2019, pp. 38--44.

\bibitem{zheng2024texttt}
R.~Zheng, X.~Wang, Y.~Sun, S.~Ma, J.~Zhao, H.~Xu, H.~Daum{\'e}~III, and F.~Huang, ``Taco: Temporal latent action-driven contrastive loss for visual reinforcement learning,'' \emph{Advances in Neural Information Processing Systems}, vol.~36, 2024.

\bibitem{yu2022structure}
J.~Yu, T.~Xu, Y.~Rong, J.~Huang, and R.~He, ``Structure-aware conditional variational auto-encoder for constrained molecule optimization,'' \emph{Pattern Recognition}, vol. 126, p. 108581, Jun 2022.

\bibitem{enoka2016translating}
R.~M. Enoka and J.~Duchateau, ``Translating fatigue to human performance,'' \emph{Medicine and science in sports and exercise}, vol.~48, no.~11, p. 2228, 2016.

\bibitem{hao2021hierarchical}
Q.~Hao, F.~Xu, L.~Chen, P.~Hui, and Y.~Li, ``Hierarchical reinforcement learning for scarce medical resource allocation with imperfect information,'' \emph{Proceedings of the ACM on Human-Computer Interaction}, 2021.

\bibitem{bottcher2008frobenius}
A.~B{\"o}ttcher and D.~Wenzel, ``The frobenius norm and the commutator,'' \emph{Linear algebra and its applications}, vol. 429, no. 8-9, pp. 1864--1885, 2008.

\bibitem{kingma2013auto}
\BIBentryALTinterwordspacing
D.~P. Kingma and M.~Welling, ``Auto-encoding variational bayes,'' Dec 2013, arXiv preprint arXiv:1312.6114. [Online]. Available: \url{https://arxiv.org/abs/1312.6114}
\BIBentrySTDinterwordspacing

\bibitem{allshire2021laser}
A.~Allshire, R.~Martin-Martin, C.~Lin, S.~Manuel, S.~Savarese, and A.~Garg, ``Laser: Learning a latent action space for efficient reinforcement learning,'' in \emph{2021 IEEE International Conference on Robotics and Automation (ICRA)}.\hskip 1em plus 0.5em minus 0.4em\relax IEEE, 2021, pp. 4951--4958.

\bibitem{watson2017informing}
M.~Watson, C.~Rusnock, M.~Miller, and J.~Colombi, ``Informing system design using human performance modeling,'' \emph{Systems Engineering}, vol.~20, no.~2, pp. 173--187, 2017.

\end{thebibliography}
\bibliographystyle{IEEEtran}
\end{document}